\documentclass[a4paper, 10 pt]{article}

\usepackage[pdftex]{graphicx}
\usepackage[tight]{subfigure}
\usepackage{algorithmic}
\usepackage{algorithm}
\usepackage{amssymb,amsmath}
\usepackage{tikz}
\usepackage{pgfplots}
\usepackage[margin=3.0cm]{geometry}

\usepackage{url}

\DeclareMathOperator*{\argmin}{arg\,min}

\title{\LARGE \bf Unsupervised Watertight Mesh Generation\\for Physics Simulation Applications\\Using Growing Neural Gas\\on Noisy Free-Form Object Models}

\author{Tobias Fromm, Christian A. Mueller, and Andreas Birk %
\thanks{Tobias Fromm and Christian A. Mueller contributed equally to this work. All authors are with the Robotics Group, Computer Science \& Electrical Engineering, Jacobs University Bremen, Germany; \emph{[t.fromm,chr.mueller]@jacobs-university.de}. This research has received funding from the European Union's Seventh Framework Programme (EU FP7 ICT-2) within the project ``Cognitive Robot for the Automation of Logistic Processes (RobLog)''.}
}
\begin{document}
\maketitle
\begin{abstract}
We present a framework to generate watertight mesh representations in an unsupervised manner from noisy point clouds of complex, heterogeneous objects with free-form surfaces. The resulting meshes are ready to use in applications like kinematics and dynamics simulation where watertightness and fast processing are the main quality criteria. This works with no necessity of user interaction, mainly by utilizing a modified Growing Neural Gas technique for surface reconstruction combined with several post-processing steps. 

In contrast to existing methods, the proposed framework is able to cope with input point clouds generated by consumer-grade RGBD sensors and works even if the input data features large holes, e.g.\ a missing bottom which was not covered by the sensor.
Additionally, we explain a method to unsupervisedly optimize the parameters of our framework in order to improve generalization quality and, at the same time, keep the resulting meshes as coherent as possible to the original object regarding visual and geometric properties.
\end{abstract}

\section{Introduction}
Surface reconstruction from point clouds has often been addressed in computer graphics, robotics, medical computing and other fields. However, depending on sampling device (e.g. consumer-grade sensors like the Kinect) and application, input points may be too sparse and noisy for conventional methods \cite{Kazhdan2006} \cite{Hoppe1992} \cite{Carr2001} \cite{Kolluri2004} to deliver appropriately polygonized surfaces. Additionally, for many applications like physics simulations, watertight meshes (featuring no holes) and even volume meshes (in addition to the surface, featuring structures also inside the mesh) are mandatory.

We present a framework which, in contrast to existing methods, is able to deliver a watertight volume mesh for any free-form \cite{Campbell2001} object, if necessary, on the expense of a roughly approximated surface. This works not only for geometric primitives, but also for more complex, heterogeneous objects with free-form surfaces.
The proposed method bases on our previous work which deals with surface reconstruction using \emph{Growing Neural Gas~(GNG)} \cite{Mueller2012} and the registration of noisy 2.5D point clouds from different viewpoints to create a 3D object point cloud \cite{Mihalyi2015}. Although this point cloud generally misses the bottom of the objects due to self-occlusion, we employ several established methods to still create meshes usable in physics simulations.

\begin{figure}[hbtp]
  \centering
  \includegraphics[width=0.7\textwidth]{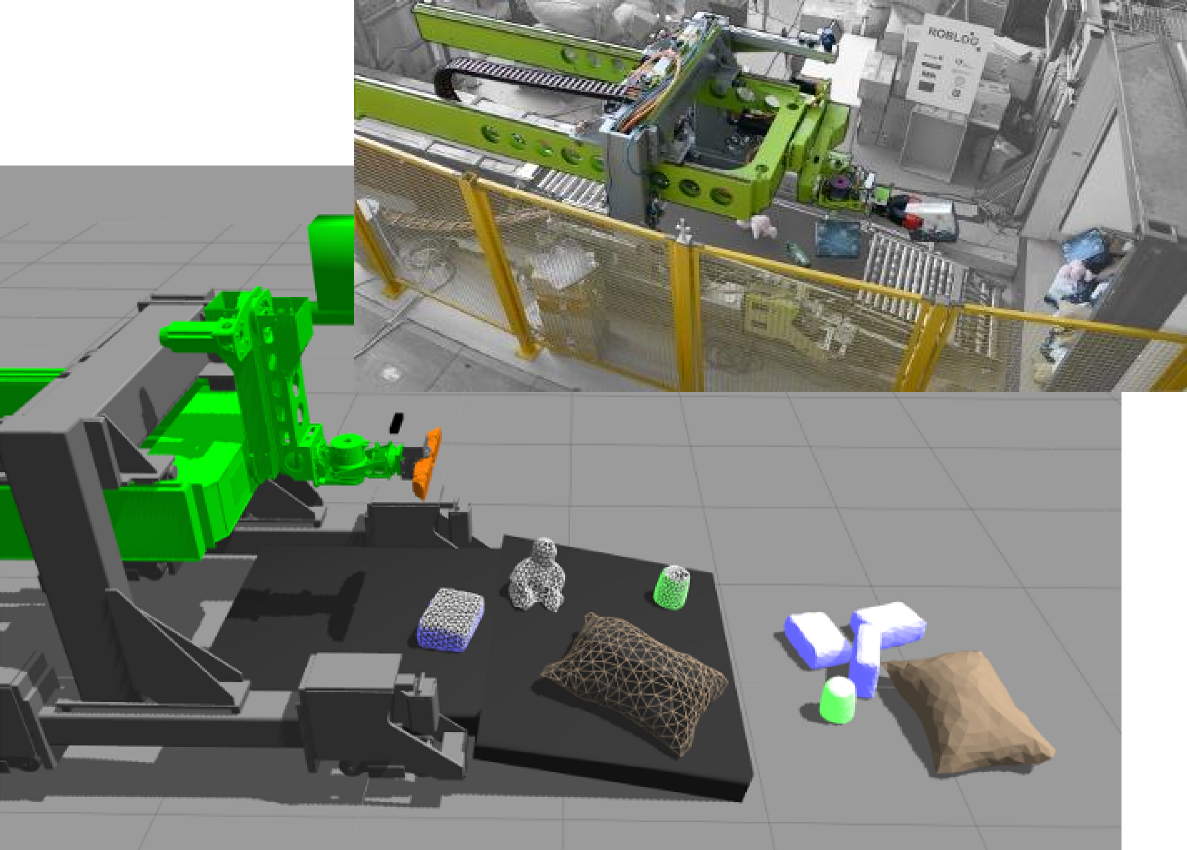}
  \caption{Application scenario: simulation of real-world object handling}
  \label{fig:roblog}
\end{figure}

The generated volume mesh representations can be used, for instance, for simulation purposes within a robotic ma\-ni\-pu\-la\-tion scenario where they are shown in a virtual scene and a virtual robot can interact with them.
Our main application lies in the EU project \emph{RobLog} with the objective to automate logistic processes, such as unloading containers \cite{Stoyanov2016}. The used hardware setup and the corresponding \emph{Gazebo} \cite{Koenig2004} physics simulation scene are shown in Fig.~\ref{fig:roblog}.
Due to the nature of physics engines, watertightness is an essential requirement which prevents models from interpenetrating.

Additionally, in order to improve the generalization performance as well as the accuracy of our approach, we present a method to optimize its parameters in an unsupervised manner using \emph{Particle Swarm Optimization (PSO)} \cite{Kennedy1995}.

\textbf{The dataset we use in this paper is publicly available\footnote{http://robotics.jacobs-university.de/datasets/2016-watertight-meshes-v01}} and provides raw sensor data, registered point clouds and the generated meshes along with Gazebo models.

\textbf{The contribution of our work is:}
\begin{itemize}
	\item a framework which requires no user-provided parameters and no user interaction that
	\item combines several established methods to generate watertight object meshes ready to use in physics simulations
	\item from noisy RGBD point clouds of complex objects
	\item including an unsupervised technique to optimize essential parameters.
\end{itemize}
To the authors' best knowledge, our framework is the first one of its kind regarding the entirety of the above criteria.

\section{Related Work}\label{sec:rw}

As for surface reconstruction, several methods exist which have been used frequently on different kinds of reconstruction problems. We use them as a baseline for our method, as described in Section~\ref{sec:comparison}.

One class of these methods bases on determining a function of the surface given by the input point set and subsequently creating polygons from the surface function using Marching Cubes \cite{Lorensen1987}.
The first one of these is Hoppe et al.'s method \cite{Hoppe1992} which estimates a signed distance function and determines its zero set. Secondly, Carr et al.'s approach \cite{Carr2001} uses a \emph{Radial Basis Function (RBF)} to describe the input data and again polygonizes the result via Marching Cubes.

Equally well-known is the method of Kazhdan et al.\ \cite{Kazhdan2006} which is based on optimizing Poisson equations for approximating a surface. This works similarly to using RBFs, but adds hierarchies in the equations which allows for locally fine-grained reconstruction where necessary. On the other hand, RBFs, working globally on the input data, may prove impractical on input data like ours with varying density. 

Another method we took as a baseline was Kolluri et al.'s \emph{Eigencrust} \cite{Kolluri2004} for which its authors emphasize its particular noise-robustness. It uses a Delaunay tetrahedralization of the input points of which each tetrahedron is labelled as \emph{inside} or \emph{outside} the object. Subsequently, the triangles which intersect between either set are used as the requested surface. Their main contribution in comparison to previous approaches is the usage of a spectral partitioning technique to divide the point sets which they claim to be more effective in distinguishing the object's edges and the environment.

Additionally, we already extensively evaluated \emph{KinectFusion} \cite{Newcombe2011} in our previous work \cite{Mihalyi2015}. Hence, it will not be taken into account again in this paper since it has already been discussed in detail that, due to some deficiencies, \emph{KinectFusion} do not provide models of the needed quality.

In the context of our application, we found the mentioned methods to deliver insufficient results, as we will explain in detail in the following sections. For this reason, the work described in this paper uses \emph{Growing Neural Gas} which interprets surface reconstruction as a learning problem.
Growing Neural Gas which was proposed by Fritzke~\cite{Fritzke1995} is an unsupervised learning technique that can be applied to learn the topology of a given distribution.
The learned topology is reflected by a graphical representation consisting of vertices (so-called neurons) and edges.
We exploit GNG learning to learn the topology of a given point cloud which can be seen as a three-dimensional distribution.
The resulting graph can be interpreted as a surface mesh and used as such for further processing steps.
In our previous work~\cite{Mueller2012} we adapted the original GNG approach~\cite{Fritzke1995} to surface reconstruction-specific properties.
Using GNG for surface reconstruction shows several advantages as described in Section \ref{sec:ng}, amongst others, \emph{online learning}, which is the ability to stop the reconstruction process any time while always receiving a triangulated surface.

Online learning has, for instance, been used in recent work of Vierjahn and Hinrichs \cite{Vierjahn2015} which introduces a reconstruction method comparable to our approach, but with no guarantee for watertightness which is a crucial requirement in our desired application scenario.

Growing Neural Gas relies on some parameters which change its behavior in several ways. Leaving these parameters on default values as given by Fritzke~\cite{Fritzke1995} and our previous work \cite{Mueller2012} yields non-optimal results because GNG is not able to adapt its topology to the desired granularity due to different requirements defined in their work.
In contrast to other approaches which deal with surface reconstruction and use a fixed, heuristically-determined parameter set \cite{Vierjahn2015} \cite{Lu2005} \cite{Marton2009}, we provide a methodology to determine the quality of the generated GNG representation and to optimize its parameters in an unsupervised manner. 
In previous work about Growing Neural Gas, this problem has not been tackled yet either \cite{Holdstein2008} or motivated in a very specific environment where it is hard to predict its general behavior \cite{GarciaRodriguez2012}.

Since the GNG-generated meshes from our method still feature some deficiencies, like missing triangulation if holes in the input data exceed a certain size, we propose several post-processing steps, among others, hole filling.
Thorough investigation has been performed on this topic already by many authors \cite{Borodin2002} \cite{Liepa2003} \cite{Ju2004}, so we decided to apply a shown-to-work method, namely \emph{MeshFix} \cite{Attene2010}, instead of displacing our focus in this direction.

\begin{figure}[tbp]
\centering  
\includegraphics[height=2.9cm]{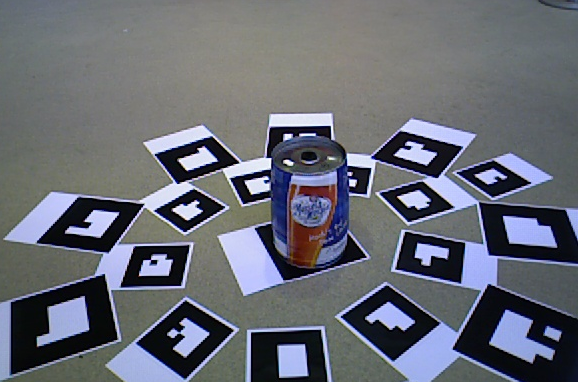}
\includegraphics[height=2.9cm]{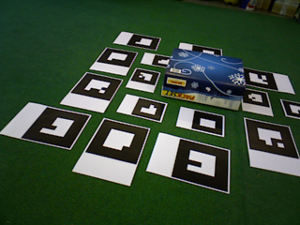}
\caption{Object modeling setup used in our previous work \cite{Mihalyi2015}}
\label{fig:modeling}
\end{figure}

Our framework, similar to all other surface reconstruction methods mentioned above, relies on multi-view 3D data. Since we want to create watertight models which cannot be penetrated by other objects, a full 3D model is required. 2.5D point clouds, as obtained from a 3D sensor from one single viewpoint, provide an incomplete model which misses (self\hbox{-})occluded parts compared to the original object's volume.
We therefore made use of our previously developed fully-automatic method \cite{Mihalyi2015} which can create 3D models without sophisticated equipment, solely using a Kinect-like RGBD sensor and printed Augmented Reality (AR) markers like in Fig.~\ref{fig:modeling}.
After capturing a number of single-view shots while manually moving the sensor around the object, registration and pose-refinement are performed.

One important feature of our previous work \cite{Mihalyi2015} is its insensitivity to variance in environment conditions, e.g.\ regarding the lighting.
Because of its ease of use even for untrained persons, this method fits the environment of the approach presented in this paper where unsupervised processing with no human intervention is desired.

\section{Watertight Volume Mesh Generation}\label{sec:method}

In Alg.~\ref{alg:method} we provide an overview of the major processing steps of our method after which we will explain every step in detail. 
In each line in Alg.~\ref{alg:method} a reference is given where to find further information beyond of what will be described in the following paragraphs.

\begin{algorithm}
\centering
\caption{Mesh Generation}
\label{alg:method}                       
\begin{algorithmic}[1]                   
\STATE \textbf{input}: noisy 3D object point cloud, e.g., from \cite{Mihalyi2015}
\STATE reconstruct surface using GNG
($\rightarrow$ Alg.\ \ref{alg:GNG_train})
\STATE remove close-by edges ($\rightarrow$ Alg.\ \ref{alg:close})
\STATE fill holes using \emph{Meshfix} \cite{Attene2010}
\STATE remove duplicate vertices/faces using \emph{VCG Library} \cite{Cignoni2014}
\STATE \emph{optional}: simplify mesh using \emph{QSlim} \cite{Garland1997}
\STATE \emph{optional}: tetrahedralize mesh using \emph{TetGen} \cite{Si2005}
\STATE restore color information ($\rightarrow$ Alg.\ \ref{alg:color})
\STATE \textbf{output}: object volume mesh
\end{algorithmic}
\end{algorithm}

\subsection{Input Data Quality Requirements}\label{sec:input}

\begin{figure*}[tb]
  \centering
  \subfigure[input cloud (top view)]{
    \label{fig:box1}
    \includegraphics[width=0.25\textwidth]{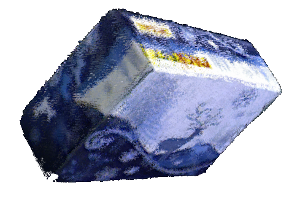} 
  }
  \subfigure[watertight mesh (top view)]{
    \label{fig:box2}
    \includegraphics[width=0.25\textwidth]{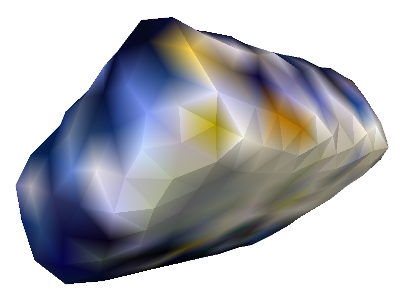} 
  }
  \subfigure[watertight mesh wireframe (top view)]{
    \label{fig:box_wireframe}
    \includegraphics[width=0.25\textwidth]{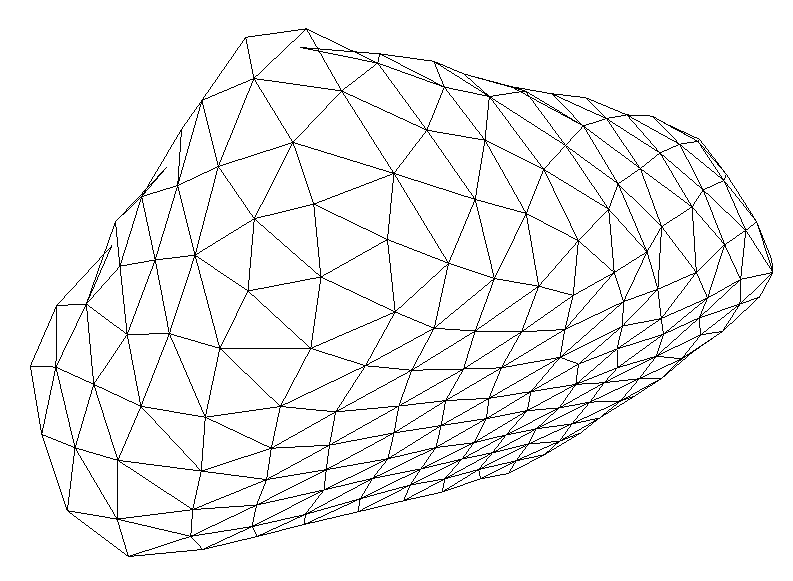} 
  }
  \subfigure[input cloud (bottom view, note the open bottom)]{
    \label{fig:box3}
    \includegraphics[width=0.25\textwidth]{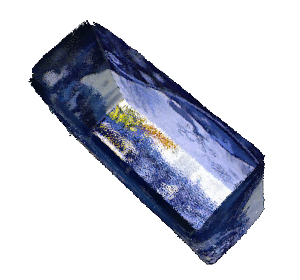} 
  }
  \subfigure[watertight mesh (bottom view, hole has been closed)]{
    \label{fig:box4}
    \includegraphics[width=0.25\textwidth]{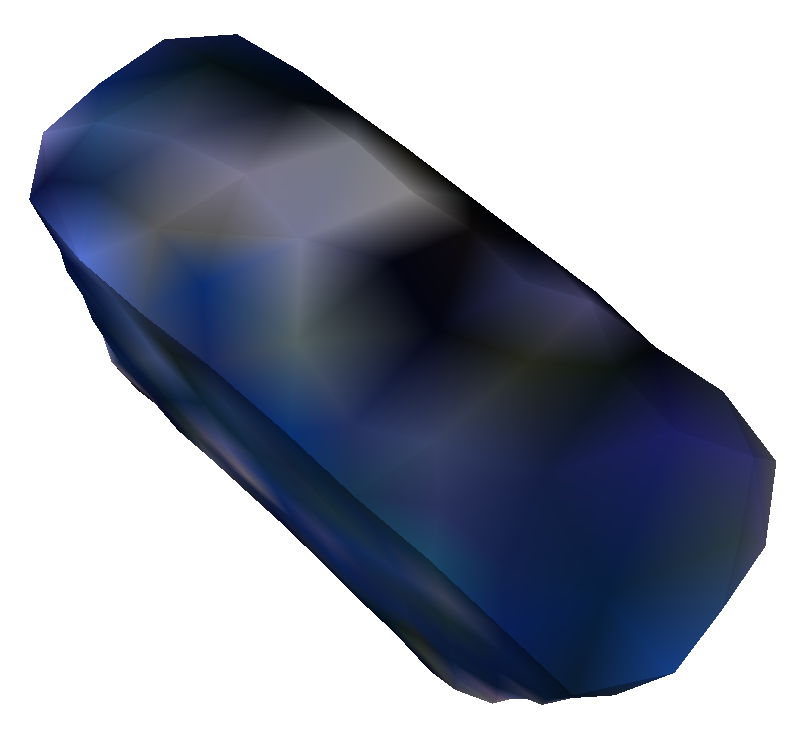} 
  }
  \caption{Watertight mesh generated via the proposed framework with no user interaction}
  \label{fig:box}
\end{figure*}

In the presented method, we use object models generated with a Kinect-like low-cost RGBD sensor and AR markers from our previous work \cite{Mihalyi2015}.
These point cloud models consist of a magnitude of 500k-1.5M points (750k in the example in Fig.~\ref{fig:box}) including the noise stemming from the consumer-grade sensor and its movement while recording the data for object modeling.

\subsection{Surface reconstruction with Growing Neural Gas}\label{sec:ng}
Growing Neural Gas is an unsupervised learning technique which aims to learn the topology of a given input distribution.
In an iterative manner using competitive Hebbian Learning \cite{Fritzke1995}, the topology of the distribution is reflected in a graphical model.
Basically, in each iteration a random sample point is selected from the distribution and fired into the graphical model.
GNG aims to adapt the graphical model by modifying the number of vertices, edges, and poses of vertices such that the input distribution is optimally approximated.
Note that a vertex in the model does not represent a sample point from the input distribution, but a prominent location in the distribution.
Since the model is extended during each iteration, this represents an online learning algorithm which can be stopped anytime while always generating a reasonable output distribution.

In case of a three-dimensional input distribution, which in our case is the input point cloud, the finally generated graphical model can be interpreted as a surface mesh.
Several beneficial properties can be observed: Due to the iterative manner of adaptation to a three-dimensional input distribution the modified GNG procedure \emph{reorganizes} vertices and edges such that prominent geometric properties are reflected in the resulting mesh.
As a consequence, the GNG results in a \emph{denoised} surface mesh -- outliers in the input distribution will have a minor or even negligible effect on the surface mesh.  
Moreover it can be shown that if a sufficient number of iterations is applied, the resulting surface mesh will converge to a \emph{Delaunay-triangulated} mesh~\cite{Fritzke1995}.

In our previous work~\cite{Mueller2012} we adapted the original GNG approach~\cite{Fritzke1995} to the specifics of the surface reconstruction problem.
A set of modifications was introduced: (a) instead of adding new vertices in a fixed interval during the learning iterations, we add a new vertex to the model $\mathcal{G}$ only if the accumulated error ($\Delta$error$(\cdot)$ -- cf.~\cite{Fritzke1995}) of a vertex in $\mathcal{G}$ exceeds a threshold $t_\gamma$; (b) we remove diverted vertices which are not coherent with the input distribution; (c) we add mild noise to the input distribution to enhance the triangulation process during training; (d) we perform multiple epochs of retraining on the model to gain a more coherent model relative to the input distribution.

For the last modification an additional convergence criterion ${\text{\emph{is\_converged}}}(\cdot)$ (Eq.~\ref{eq:gng_conver_crit}) was introduced to analyze the consistency between the learned GNG model $\mathcal{G}_{t}$ for training epoch $t$ and the input distribution (point cloud) $\mathcal{P}$,
\begin{equation} \label{eq:gng_conver_crit}
\text{\emph{is\_converged}}(\mathcal{G}_{t}) = 
\begin{cases}
        \begin{array}{lcl}
        \text{\emph{true}} \text{ if } \varepsilon(\mathcal{G}_{t}) > \varepsilon(\mathcal{G}_{t-1}) \\
        \text{\emph{false}} \text{ otherwise}
        \end{array}
\end{cases}
\end{equation}
\begin{equation} \label{eq:gng_consist}
    \varepsilon(\mathcal{G}_{t}) = \frac{\displaystyle\sum_{v \in \mathcal{G}} \argmin_{p \in \mathcal{P}} (\|v-p\|_{2})}{|\mathcal{G}_{t}|}
\end{equation}

where $\varepsilon(\mathcal{G}_{t})$ is a score which, based on the \emph{$L^2$-norm}, computes the consistency between $\mathcal{G}_{t}$ and $\mathcal{P}$, $|\mathcal{G}_{t}|$ represents the number of vertices in $\mathcal{G}_{t}$, $v$ denotes a vertex from $\mathcal{G}_{t}$ and $p$ a point in $\mathcal{P}$.
After each retraining epoch we compute $\text{\emph{is\_converged}}(\mathcal{G}_{t})$ for the current state of the retrained GNG model. 
If $\text{\emph{is\_converged}}(\mathcal{G}_{t})$ returns \emph{true} the training will be stopped and $\mathcal{G}_{t}$ will represent the final surface mesh for the given point cloud $\mathcal{P}$, otherwise a new training epoch is applied on $\mathcal{G}_{t}$.

Alg.~\ref{alg:GNG_train} provides a brief overview of the epochal GNG learning. 
For further details about the Growing Neural Gas learning procedure we refer to \cite{Fritzke1995} and our previous initial work \cite{Mueller2012} on which we build here.

\begin{algorithm}
\centering
\caption{GNG-based Surface Reconstruction
}
\label{alg:GNG_train}
\begin{algorithmic}[1]
    \STATE \textbf{input}: point cloud $\mathcal{P}$
    \STATE create an empty GNG Model $\mathcal{G}_{t}$, $t=0$
    \WHILE{$\lnot \text{\emph{is\_converged}}(\mathcal{G}_{t})$}
    \STATE train GNG on $\mathcal{G}_{t}$ using $\mathcal{P}$
    \STATE $t=t+1$
    \ENDWHILE{}
    \STATE \textbf{output}: generated surface mesh $\mathcal{G}$ = $\mathcal{G}_{t}$
\end{algorithmic}
\end{algorithm}

\subsection{Post-Processing}\label{sec:pp}
After surface reconstruction, several post-processing steps are performed to make the mesh watertight, see Alg.~\ref{alg:method}.

\paragraph{Close-By Edge Removal}
Edges that are very close to each other, i.e. in the range of the point cloud's resolution, are removed in this step in order to not maintain overlapping triangles. This step, developed by the authors, is described in Alg.~\ref{alg:close} and was integrated because the GNG might create polygons of order $>3$ with connected vertices.
These edges are then very close to each other without necessarily intersecting because the vertices of the created polygons usually do not lie on a flat surface.
However, the following post-processing steps cannot deal with such contingencies in a way that a smooth surface is created regardless of the presence of overlapping and intersecting triangles, thus this step is crucial.

\begin{algorithm}
\centering
\caption{Close-By Edge Removal}
\label{alg:close}
\begin{algorithmic}[1]
\FOR {all vertices $v_{i} \in \mathcal{G}$}
    \FOR {all neighboring vertices $v_{j}$ of $v_{i}$}
        \FOR {all neighb.\ vertices $v_{k}$ of $v_{j}$ where $v_{k} \ne v_{i}$}
            \IF {$d(e_{ij},e_{jk})<t_{p}$}
                \STATE remove $e_{jk}$
            \ENDIF
        \ENDFOR
    \ENDFOR
\ENDFOR

where $e_{ij}$ is the edge between vertices $v_{i}$ and $v_{j}$, $d(e_{ij},e_{jk})$ is the Euclidean distance between the closest points of $e_{ij}$ and $e_{jk}$, $t_{p}$ is a threshold for the maximum allowed edge proximity
\end{algorithmic}
\end{algorithm}

A suitable value for the threshold $t_{p}$ in Alg.~\ref{alg:close} has to be determined experimentally, depending on the resolution of the input point clouds. The value should be as low as possible in order to not erroneously remove edges of different parts of the model, but still to remove all intersecting or close-by edges. For our input dataset, any value of $t_{p}$ in the range $[0.005,0.01]~m$ works sufficiently and removes all close-by artifact edges, but leaves the remaining edges intact.

\paragraph{Hole Filling}
Next, we use \emph{MeshFix} \cite{Attene2010} to fill the remaining holes in the mesh. Despite the GNG managing to reconstruct the surface close to the input point cloud, holes may remain due to sparse input data. The reasons for this sparsity may be reflections, illuminance issues, etc.\ which usually appear in 3D data obtained from low-cost sensors.
Additionally, the registered point cloud does not cover the bottom of the object. This leads to a hole which has to be filled.
As demonstrated in Fig.~\ref{fig:examples}, this post-processing step enables our framework to deal with big extents of self-occlusion and otherwise missing input data.

\paragraph{Duplicate Face Removal}
During the hole filling process, duplicate faces might appear. Thus, the next step is to remove them utilizing the respective function of the \emph{VCG Library} \cite{Cignoni2014} because otherwise this may cause the tetrahedralization routine (see below) to be unable to generate tetrahedra between two flatly aligned, congruent faces.

\paragraph{Simplification}
As an optional step, in case the user desires an even more sparse mesh resolution for faster processing within their application, experiments have shown that \emph{QSlim} \cite{Garland1997} provides effective mesh simplification. As shown in Fig.~\ref{fig:examples}, the generated meshes though contain triangles in a magnitude of several hundreds only and will be coarse-grain enough for sufficient processing speed, so generally this step will not be necessary for most applications.

\paragraph{Tetrahedralization}
Depending on the application, a tetrahedralization of the resulting surface meshes may be necessary to obtain volume meshes. This is crucial, for instance, for the fidelity of soft-body simulations where an object is not only defined by its boundaries, but also its intrinsic structure. In a dynamics simulation, surface-only meshes will collapse once the physics engine is stepped. Hence, we propose the use of \emph{TetGen} \cite{Si2005} which we successfully use to tetrahedralize the generated surface meshes. For pure rigid-body simulation, this step can be omitted.

\paragraph{Color Information Restauration}
Eventually, the color information needs to be restored since it was removed during the surface generation and refinement process. Not for all applications it is necessary to have colored meshes, but this is a process which can efficiently be applied. This step makes use of a $k$-d tree structure and is described in detail in Alg.~\ref{alg:color}.
\begin{algorithm}
\centering
\caption{Color Information Restauration}
\label{alg:color}
\begin{algorithmic}[1]
    \STATE \textbf{input}: reconstructed mesh $\mathcal{G}$, colored point cloud $\mathcal{P}$
    \STATE create $k$-d tree from $\mathcal{P}$
    \FOR {all $p_{i} \in \mathcal{G}$}
        \STATE search tree for (spatial) nearest neighbor of $p_{i}$
        \STATE set color of $p_{i}$ on color of nearest neighbor in $\mathcal{P}$
    \ENDFOR
\end{algorithmic}
\end{algorithm}

\section{Mesh quality optimization}\label{sec:optimization}
Growing Neural Gas requires a set of parameters to be adapted to the quality of the input data in order to create a mesh with properties considered as optimal.
On the one hand, high generalization skills during reconstruction are required for noise removal and fast processing, but the generated meshes should not be too coarse to lose details on complex objects. On the other hand, low generalization tends to overfit the input data and does not close holes efficiently. 

For these reasons, we present an approach to optimize the GNG parameters as used in Alg.~\ref{alg:method} in an unsupervised manner, using several optimization criteria.
Both the GNG parameters and mesh optimization criteria will be described in detail in the following.

\subsection{Parameters to optimize}\label{sec:parameters}
In order to achieve well-reconstructed surfaces from noisy point clouds, a set of parameters is required to be appropriately set up.
The selected GNG parameters to optimize are shown in Table~\ref{table:gng_params} where $\mathcal{P}$ denotes a three-dimensional distribution (which represents the input point cloud) and $\mathcal{G}$ a trained GNG model on $\mathcal{P}$.
These parameters depend on the dimensions of the input point cloud as well as the inherent and encompassing noise and thus need to be optimized.

\begin{table}[htbp]
\caption{Growing Neural Gas Parameters}\label{table:gng_params}
\centering
\begin{tabular}{l|l}
\textbf{parameter}&\textbf{description}\\
\hline
$\epsilon_b$& move nearest vertex towards the signal by this fraction\\\hline 
$\epsilon_n$& move all adjacent vertices of the nearest vertex to signal\\
	    & towards the signal by this fraction\\\hline 
$a_{\text{max}}$ & remove edges in model $\mathcal{G}$ which are older than $a_{\text{max}}$\\ 
                  & within a GNG training epoch\\\hline 
$t_\gamma$  & a new vertex is added only if the $\Delta$error$(\cdot)$ of a \\
            & vertex $v\in\mathcal{G}$ exceeds $t_\gamma$\\ \hline 
$\alpha$& if a new vertex is added decrease the error variable \\
& of the nearest and second nearest vertex by this factor\\\hline 
$d$& factor to decrease the error variable of each vertex\\
   & in model $\mathcal{G}$\\ 
\end{tabular}

\centering Essential parameters applied in each GNG training epoch (cf.\ \cite{Mueller2012} and \cite{Fritzke1995})
\end{table}

The procedure to optimize this 6-dimensional parameter set will be described in the following; first, we need to define the criteria with respect to which to optimize.

\subsection{Mesh optimization criteria}\label{sec:criteria}
In the following, we present two criteria we found to be relevant to have an effect on the mesh quality.

\paragraph{Consistency Error $\varepsilon(\mathcal{G})$}
defined as the consistency between $\mathcal{G}_{t}$ and $\mathcal{P}$, see Eq.~\ref{eq:gng_consist}

\paragraph{Mean Edge Length Ratio $\eta(\mathcal{G})$}
\begin{equation}
	\eta(\mathcal{G}) = 1-\frac{\sum_{1}^{n}\frac{l_{\mathrm{min}}}{l_{\mathrm{max}}}}{n}
\end{equation}
where $n$ is the number of triangles in the mesh, $l_{\mathrm{min}}$ is the length of the shortest, $l_{\mathrm{max}}$ the length of the longest edge of the respective triangle.

From these criteria, $\varepsilon(\mathcal{G})$ describes the consistency between the generated mesh and the input data where $\eta(\mathcal{G})$ correlates with the mesh surface appearance.

Both criteria are combined into an \emph{evaluation~function~$\theta(\mathcal{G})$} for each time step $t$ in the following way:
\begin{equation}
\theta(\mathcal{G}) = 
\begin{cases}
        \begin{array}{ll}
        \text{\emph{true}} & \text{if } (\varepsilon_{t+1}(\mathcal{G}) < \varepsilon_{t}(\mathcal{G}) \land \eta_{t+1}(\mathcal{G}) \leq \eta_{t}(\mathcal{G})) \\
		& \hspace{0.1cm} \lor (\varepsilon_{t+1}(\mathcal{G}) \leq \varepsilon_{t}(\mathcal{G}) \land \eta_{t+1}(\mathcal{G}) < \eta_{t}(\mathcal{G})) \\
        \text{\emph{false}} & \text{otherwise}
        \end{array}
\end{cases}
\label{eq:evaluation}
\end{equation}

Instead of the condition in Eq.~\ref{eq:evaluation}, pure \emph{and}-relations, \emph{or}-relations or a product of $\varepsilon(\mathcal{G})$ and $\eta(\mathcal{G})$ would have been possible to use.
In our experiments, we disregarded these combinations due to different numeric ranges. Especially since $\varepsilon(\mathcal{G})$ measures a distance, it has no upper boundary; typical values of $\varepsilon(\mathcal{G})$ are in a $10^{-3}$ magnitude whereas $\eta(\mathcal{G})$ is a ratio of two equivalent values and thus normalized on $[0,1[$.
As for \emph{and} and \emph{or}-relations, one criterion is favored on the expense of the other in case of divergence, thus this will lead to a perturbed optimization.

Another possible mesh optimization criterion is a Hausdorff-like distance between input point clouds and meshes which was proposed to measure surface distances in several works \cite{Cignoni1998} \cite{Aspert2002}.
However, the processing intensity of this measure (an all-to-all comparison of input points and mesh vertices) makes it impractical to use in our case because it increases our framework's runtimes about 10-fold.

\subsection{Particle Swarm Optimization}
Many existing approaches use hand-crafted values for their parameters to achieve satisfactory results. Moreover, parameters can form strong relationships between each other.

Since the parameter optimization problem presented here is discontinuous, all optimization methods relying on a continuous function describing the problem will fail.
As for pattern search-based methods, the brute-force Grid Search is one of the most trivial, but least efficient ones. Instead, we utilize \emph{Particle Swarm Optimization (PSO)} \cite{Kennedy1995} in a parallelized implementation of \cite{Kennedy2001}. 

PSO can be used without assumptions about the data to optimize. As an input, it is able to deal with a number of numerical parameters in arbitrary ranges (for our case, see Table~\ref{table:gng_params}) and needs an evaluation function towards which to optimize ($\theta(\mathcal{G})$).
It will first create a random configuration within the 6-dimensional space formed by the GNG parameters and randomly distribute a swarm of agents around it.

As shown in Alg.~\ref{alg:pso}, with every step, the GNG is trained and the evaluation function $\theta(\mathcal{G})$ is resolved.
Afterwards, the center of the agent distribution shifts in the direction of the current global optimum.
After a specified maximum number of iterations, the position of the globally optimal agent determines the optimized set of GNG parameters to be used from then on.

\begin{algorithm}
\centering
\caption{Particle Swarm Optimization}
\label{alg:pso}
\begin{algorithmic}[1]
    \STATE \textbf{input}: dimensions of parameter space $\mathcal{A}$
    \STATE initialize all agents $a_{i} \in \mathcal{A}$ randomly
    \STATE initialize globally optimal agent $a_{\mathrm{opt}} = a_{0}$
    \WHILE {less than maximum number of iterations}
	\FOR {all $a_{i} \in \mathcal{A}$}
	        \STATE train GNG using $a_{i}$
		\IF {$\theta(\mathcal{G})$}
		        \STATE update optimal agent $a_{\mathrm{opt}} = a_{i}$
		\ENDIF
	\ENDFOR
	\STATE update all agents' parameter sets to move the swarm towards the current optimum
    \ENDWHILE
    \STATE \textbf{output}: globally optimal agent $a_{\mathrm{opt}}$
\end{algorithmic}
\end{algorithm}

We used the default settings for PSO as they yielded satisfying results in our experiments (see Section~\ref{sec:pso}). Meta-optimizing the Particle Swarm Optimizer has been subject to research (e.g.\ \cite{Carlisle2001}), but since the default parameters given by Kennedy and Eberhart~\cite{Kennedy1995} provide sufficient optimization convergence for our use case, we regard this topic as out-of-scope for the work proposed in this paper.

\section{Results}\label{sec:results}

\subsection{Growing Neural Gas parameter optimization}\label{sec:pso}

\begin{figure}[tbp]
\centering
\subfigure[synthetic point cloud used for optimization]{\includegraphics[width=.22\textwidth]{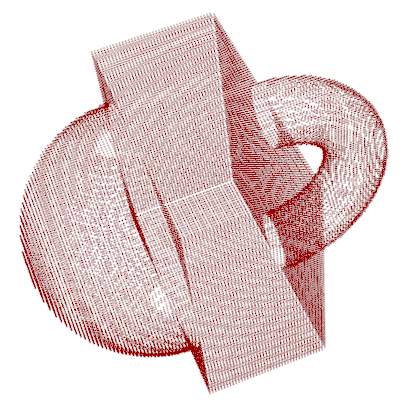}\label{fig:pso1}}
\subfigure[mesh for initial non-optimized parameters]{\includegraphics[width=.22\textwidth]{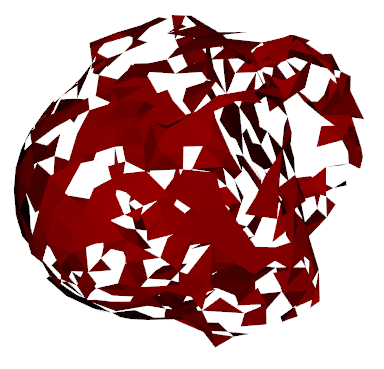}\label{fig:pso2}}
\subfigure[mesh for optimized parameters (view 1)]{\includegraphics[width=.22\textwidth]{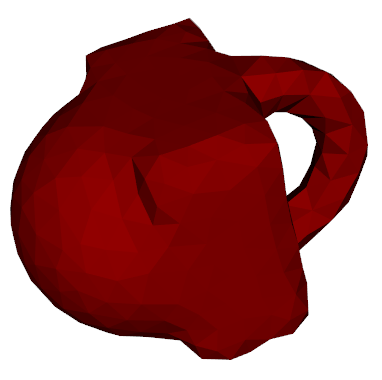}\label{fig:pso3}}
\subfigure[mesh for optimized parameters (view 2)]{\includegraphics[width=.22\textwidth]{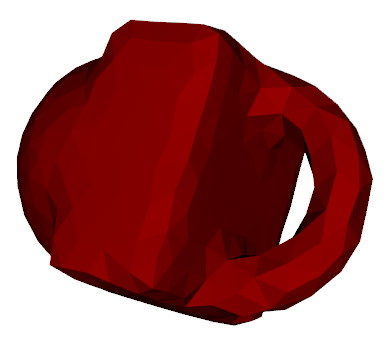}\label{fig:pso4}}
\caption{Particle Swarm Optimization results (visual)}
\label{fig:pso}
\end{figure}

For optimizing the GNG parameters, we introduce a synthetic object model shown in Fig.~\ref{fig:pso}.
The rationale behind this is to avoid model-specific bias from our object models. It combines surface features of prototype shapes like sphere, cuboid and torus, containing edges, flat and curved surfaces.
This model was intentionally created noise-free because the GNG deals with noise occuring in our real scenario.

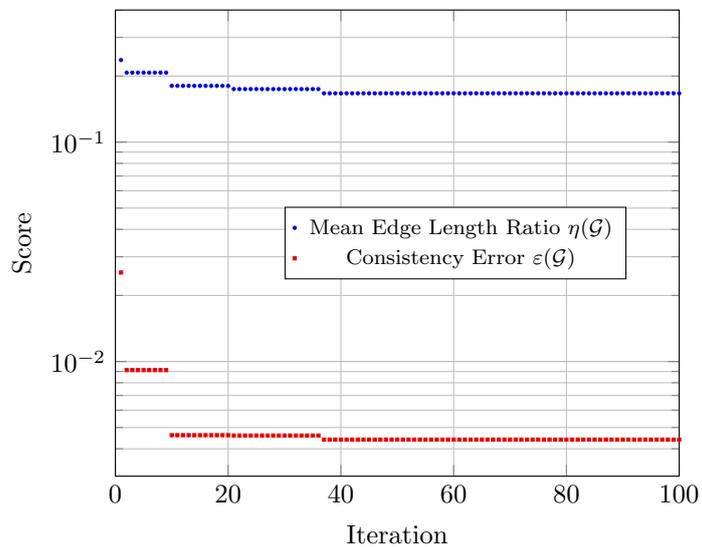
\begin{figure}[tbp]
\centering
\begin{tikzpicture}
\begin{semilogyaxis}[xlabel=Iteration, ylabel=Score, grid=both, legend style={ at={(0.3,0.5)}, anchor=west }, ymin=0.003, ymax=0.4, xmin=0, xmax=100, width=0.6\textwidth]
\addplot+[only marks,mark size=0.6pt]file{plot_consistency_error_all.csv};
\addlegendentry{\footnotesize{~Mean Edge Length Ratio $\eta(\mathcal{G})$}};
\addplot+[only marks,mark size=0.6pt]file{plot_mean_edge_length_ratio_all.csv};
\addlegendentry{\footnotesize{~Consistency Error $\varepsilon(\mathcal{G})$}};
\end{semilogyaxis}
\end{tikzpicture}
\caption{PSO results using $\theta(\mathcal{G})$ as an evaluation function}
\label{fig:pso_numeric_results}
\end{figure}

PSO was run for 100 iterations using the evaluation function $\theta(\mathcal{G})$, which combines $\eta(\mathcal{G})$ and $\varepsilon(\mathcal{G})$, with the results in Fig.~\ref{fig:pso_numeric_results}.
It can be observed that the $\eta(\mathcal{G})$ and $\varepsilon(\mathcal{G})$ scores both improve over time and converge after about 35 iterations, hence this leads to the optimized GNG parameter set given in Table~\ref{table:pso_results}.

\begin{table}[bp]
\caption{Particle Swarm Optimization results}\label{table:pso_results}
\centering
\begin{tabular}{l|l|l||l}
\textbf{parameter}&\textbf{range}&\textbf{seed}&\textbf{optimized value}\\
\hline
	$\epsilon_b$& $[0,1]$ & 0.2 & 0.0739138 \\\hline 
	$\epsilon_n$& $[0,0.2]$ & 0.006 & 0.00870156 \\\hline 
	$t_\gamma$ & $[0,5]$ & 3.0 & 2.72645 \\\hline 
	$a_{\text{max}}$ & $[50,250]$ & 60 & 133 \\\hline 
	$\alpha$& $[0,1]$ & 0.5 & 0.521687 \\\hline 
	$d$& $[0,1]$ & 0.995 & 0.999321 
\end{tabular}
\end{table}

The respective ranges were pre-set as an initial estimate, yielding faster optimization and reducing the likelihood to fall into local minima which may occur for PSO, especially for discontinous evaluation functions like $\theta(\mathcal{G})$. 
In order to accelerate the optimization process we initialize the Particle Swarm with seed parameters (cf. Table~\ref{table:pso_results}) which are based on the original parameters evaluated by Fritzke \cite{Fritzke1995}. These parameters provide a reasonable guess so that the GNG algorithm is capable to evolve an initial set of vertices.

\subsection{Comparison to similar approaches}\label{sec:comparison}
For qualitatively evaluating our framework, we used \emph{Poisson} \cite{Kazhdan2006}, \emph{Eigencrust} \cite{Kolluri2004}, as well as Hoppe et al.'s \cite{Hoppe1992} and Carr et al.'s \cite{Carr2001} methods for baselines which replaced Line~2 in Alg.~\ref{alg:method} while not changing the overall process.
All of the mentioned methods were evaluated using different objects from our testing set, see the examples in Fig.~\ref{fig:examples}.

\begin{figure}[tbp]
\centering
\subfigure[Poisson reconstruction]{\includegraphics[width=.25\textwidth]{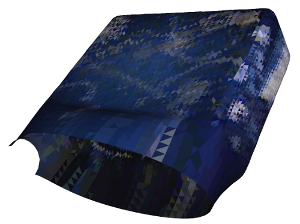}\label{fig:poisson2}}
\hspace*{0.5cm}
\subfigure[after post-processing]{\includegraphics[width=.25\textwidth]{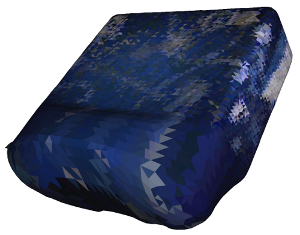}\label{fig:poisson3}}
\caption{Poisson surface reconstruction results (cf.\ our results, Fig.~\ref{fig:box})}
\label{fig:poisson}
\end{figure}

As for the reconstruction proximity to the input data, Poisson surface reconstruction delivers visually satisfactory results like in Fig.~\ref{fig:poisson2} when used with hand-tuned parameters (\emph{maximum tree depth: 6, minimum number of samples per node: 20}).
Unfortunately, the method is unable to deal with sharp object edges where it creates bulges instead of preserving the structure of the input data. The proposed post-processing does not help since it fills holes only after the original object structure has been distorted (see Fig.~\ref{fig:poisson3}). Since our input models miss at least the bottom face, this deficiency invalidates the use of Poisson surface reconstruction for our purpose.
Additionally, Poisson reconstruction sometimes creates multiple unconnected meshes for input models containing holes where GNG, by definition, generates exactly one mesh. Hence, we consider Poisson-generated meshes as visually appealing for most objects, but unsatisfactory in the context of realistic simulation.

Equally unsatisfactory is the deployment of any of the other three methods (Hoppe et al.'s and Carr et al.'s, which both make use of Marching Cubes \cite{Lorensen1987}, and Eigencrust) on our input data.
Carr et al.\ use dense models with several 100k points which are practically noise-free, Hoppe et al.'s example point sets are not very dense (several 1000), but noise-free by definition since they were sampled from CAD models. Kolluri et al.'s point clouds are noise-free as well from the low-cost sensor point of view, only bearing artifical outliers which are mostly placed far away from the object.

Concretely, Hoppe et al.'s method and \emph{Eigencrust} failed on our input data with zero generated surface vertices regardless of their parameter settings where Carr et al.'s method runs for more than one hour of CPU time without any result.

Summarized, we were unable to create a viable result with any of these methods, probably due to the fact that our input point clouds were created with a low-cost RGBD camera and hence incorporate noise these methods cannot deal with.

\subsection{Results for objects of different complexity}\label{sec:results_complexity}
\begin{figure*}[btp]
\centering
\subfigure[Barrel, $\eta(\mathcal{G})=0.209718$, $\varepsilon(\mathcal{G})=7.20993\cdot 10^{-4}$]{\includegraphics[width=.20\linewidth]{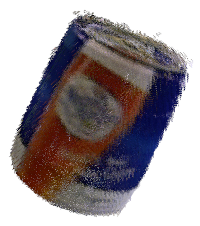}\includegraphics[width=.20\linewidth]{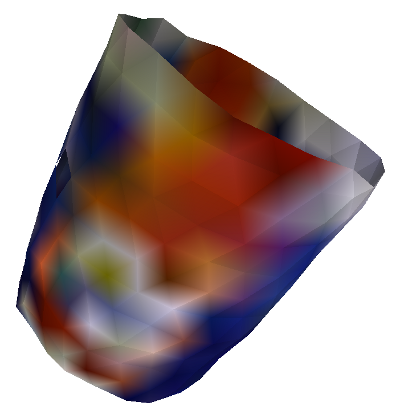}\includegraphics[width=.20\linewidth]{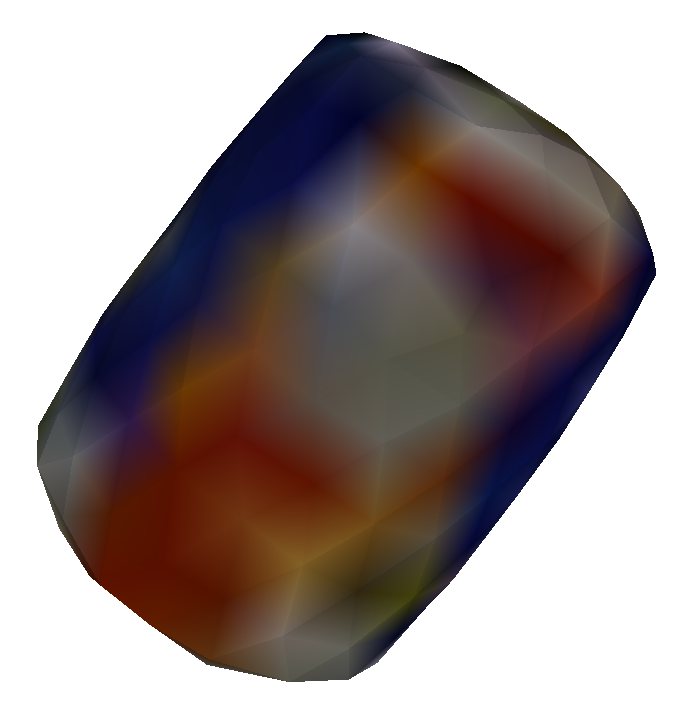}\includegraphics[width=.20\linewidth]{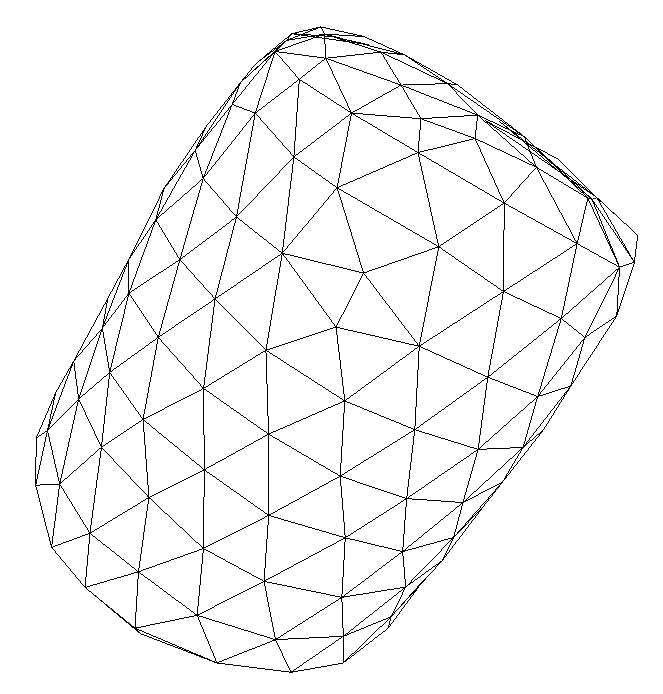}\includegraphics[width=.20\linewidth]{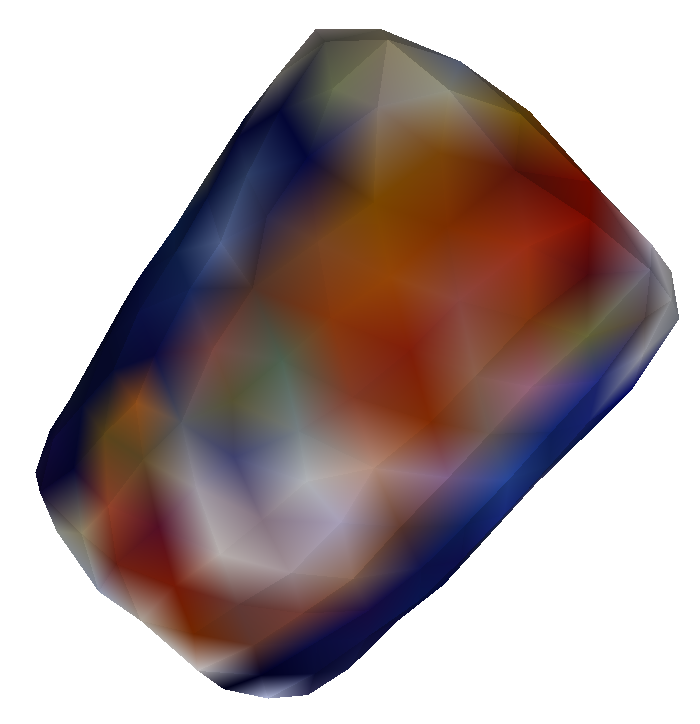}}
\subfigure[Coffee sack, $\eta(\mathcal{G})=0.204374$, $\varepsilon(\mathcal{G})=7.46082\cdot 10^{-4}$]{\includegraphics[width=.2\linewidth]{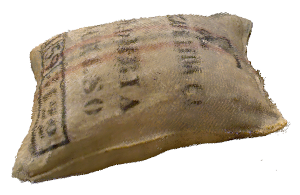}\label{fig:sack_input}\includegraphics[width=.2\linewidth]{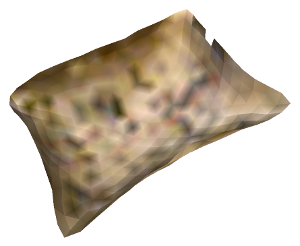}\includegraphics[width=.2\linewidth]{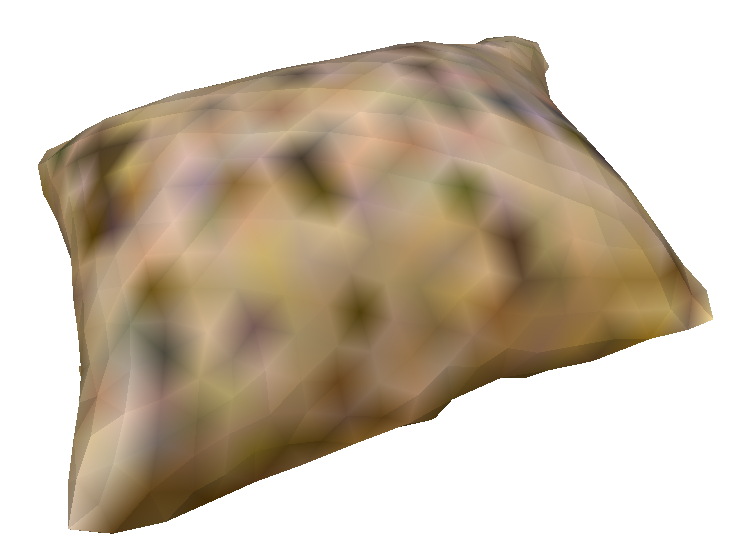}\label{fig:sack_final}\includegraphics[width=.2\linewidth]{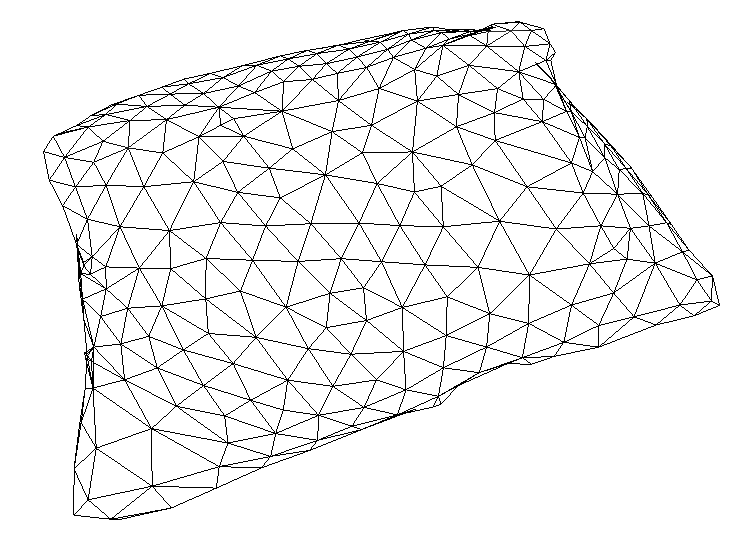}\label{fig:sack_wireframe}\includegraphics[width=.2\linewidth]{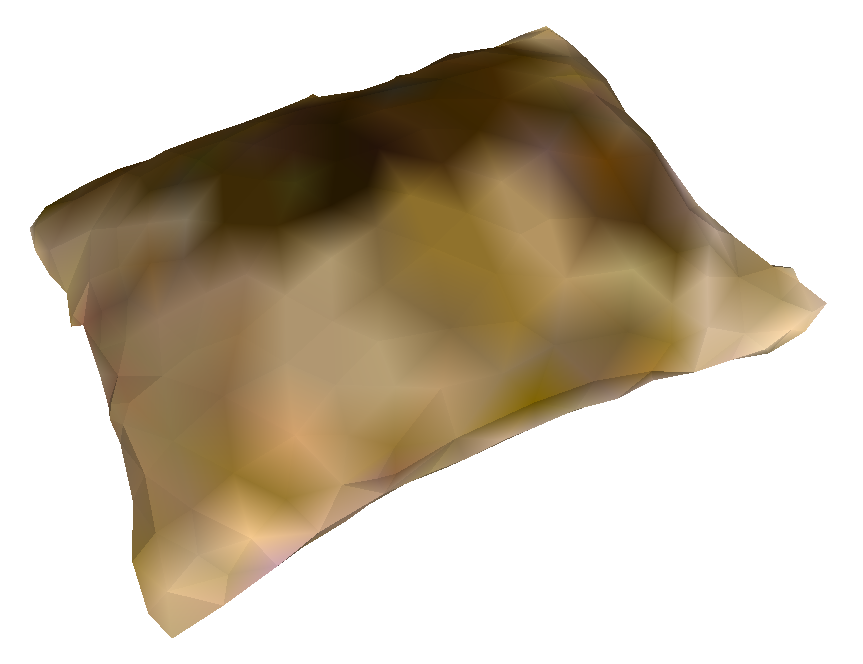}\label{fig:sack_bottom}}
\subfigure[Post box, $\eta(\mathcal{G})=0.203831$, $\varepsilon(\mathcal{G})=9.24772\cdot 10^{-4}$]{\includegraphics[width=.20\linewidth]{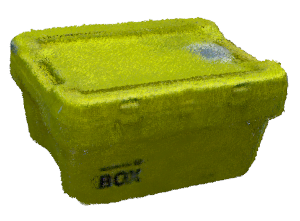}\includegraphics[width=.20\linewidth]{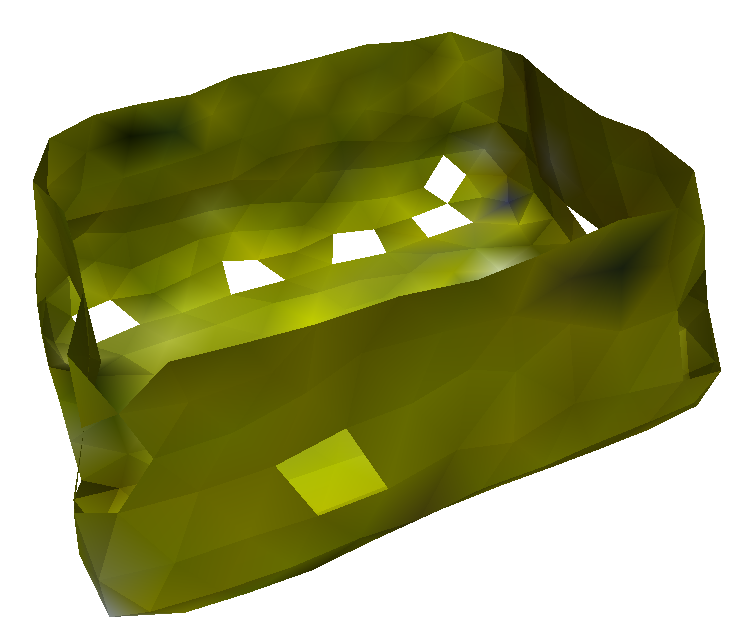}\includegraphics[width=.20\linewidth]{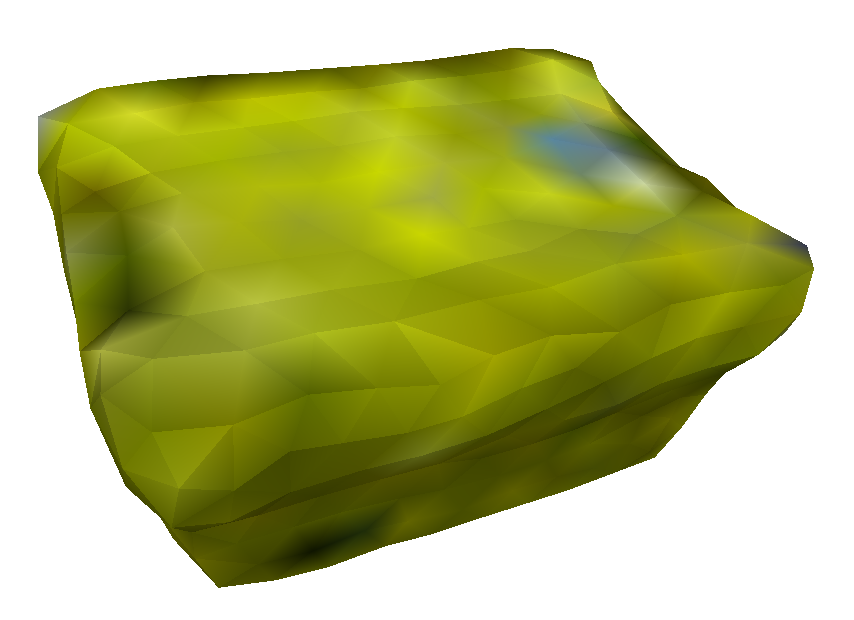}\includegraphics[width=.20\linewidth]{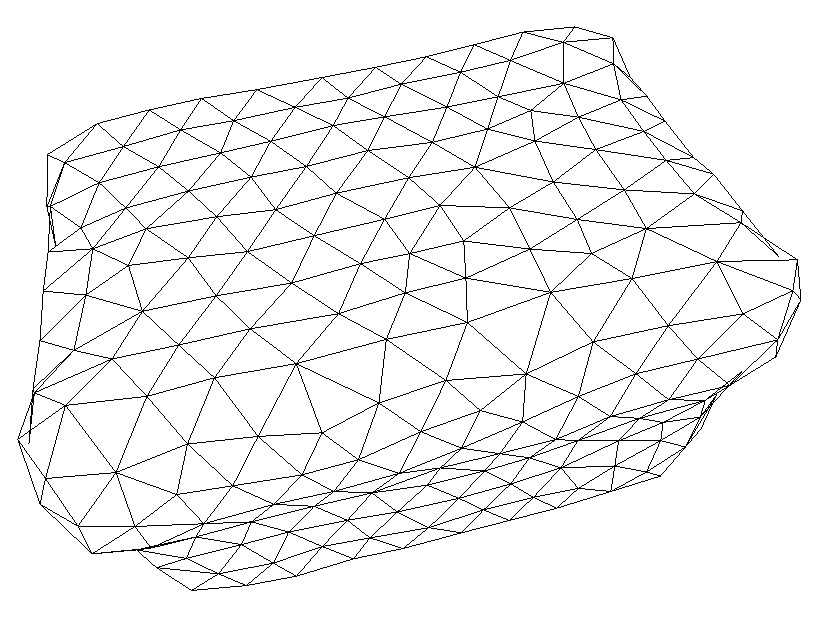}\includegraphics[width=.20\linewidth]{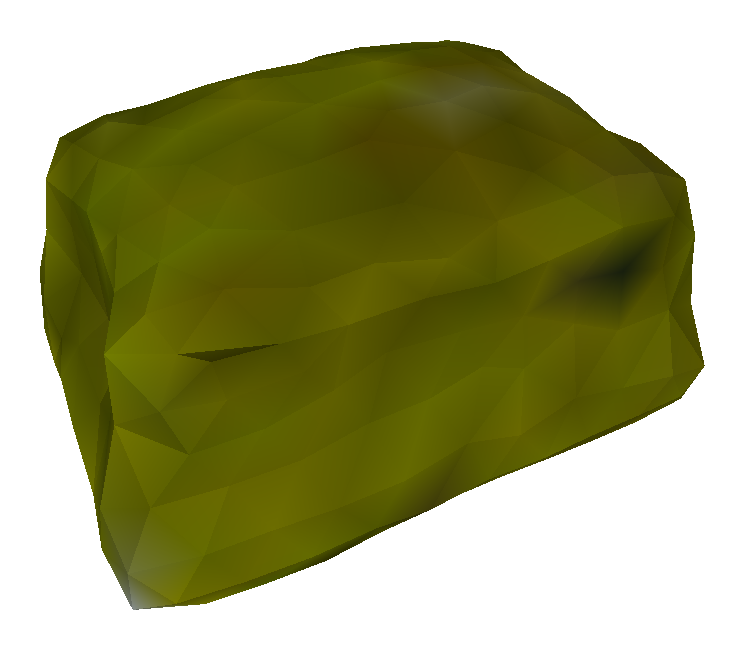}}
\subfigure[Baby doll, $\eta(\mathcal{G})=0.216012$, $\varepsilon(\mathcal{G})=8.89707\cdot 10^{-4}$]{\includegraphics[width=.20\linewidth]{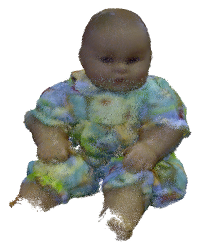}\label{fig:doll_input}\includegraphics[width=.20\linewidth]{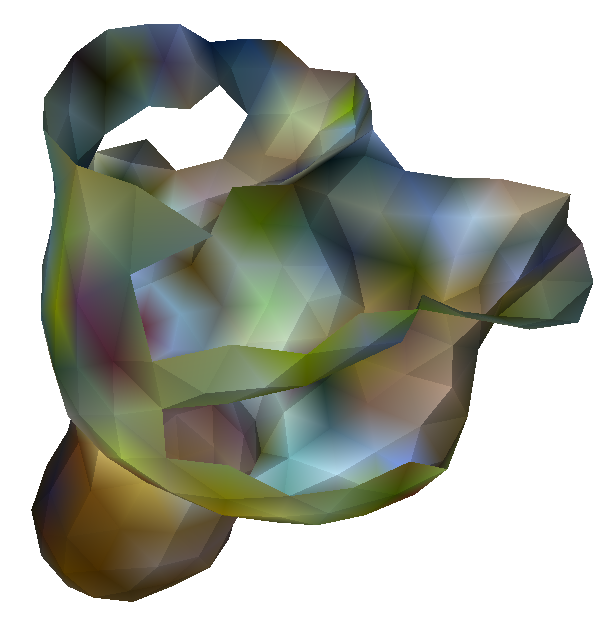}\includegraphics[width=.20\linewidth]{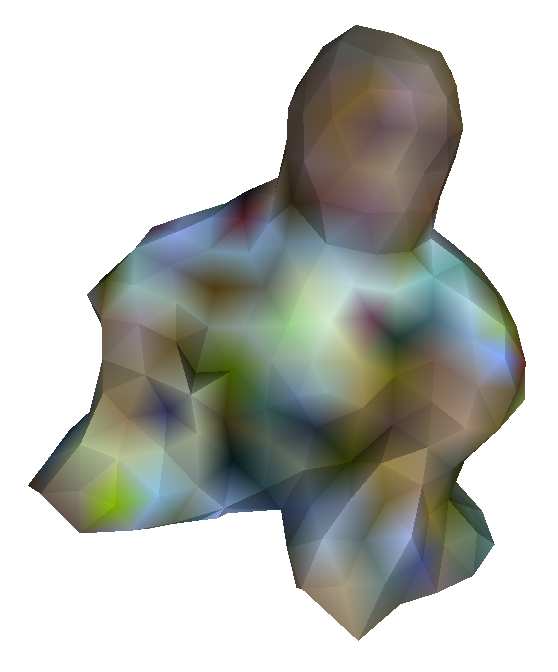}\label{fig:doll_final}\includegraphics[width=.20\linewidth]{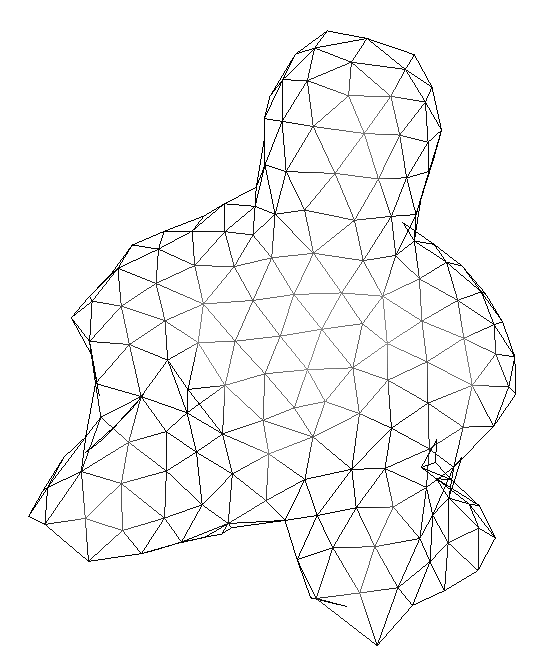}\label{fig:doll_wireframe}\includegraphics[width=.20\linewidth]{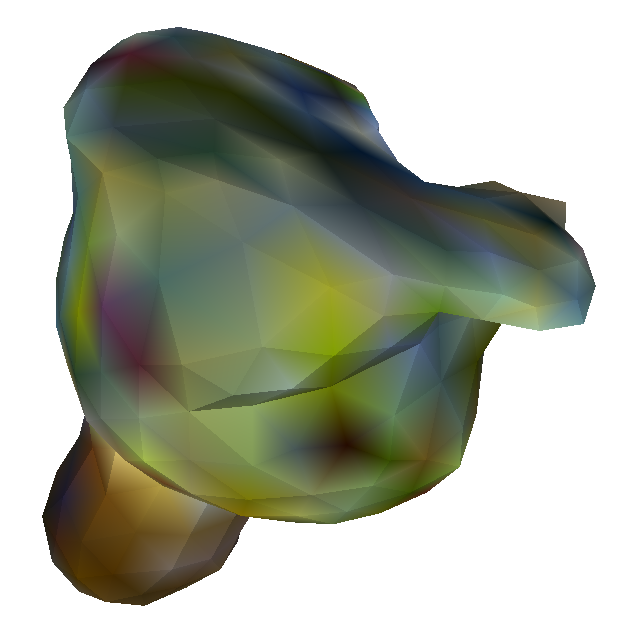}\label{fig:doll_bottom}}
\caption{Collection of object model point clouds from our previous work \cite{Mihalyi2015} and the respective meshes generated with the method proposed herein. Particle Swarm-optimized parameters were used, see Section~\ref{sec:pso}.\newline From left to right: 1. original point cloud, 2. GNG mesh (bottom view), 3. post-processed mesh (top view), 4. post-processed wireframe mesh (top view), 5. post-processed mesh (bottom view).}
\label{fig:examples}
\end{figure*}

The proposed method can not only be applied to objects which may also be modeled by geometric primitives, like the box in Fig.\ \ref{fig:box}, but also for more complex items like barrels, balls, tires, sacks and dolls. Fig.\ \ref{fig:examples} shows objects of different complexity modeled with our approach alongside with the values of their respective optimization criteria, each one featuring incomplete model information in the form of self-occlusion or missing faces. Nevertheless, our framework is capable of resolving these model deficiencies. 

\subsection{Known limitations}\label{sec:limitations}
In the following, we will identify several cases in which our method may fail. First, since any holes in the meshes will be fixed by a planar series of triangles, no additional information will be generated which is not already present in the input point cloud. 
In case of the baby doll (Fig.~\ref{fig:doll_input}--\ref{fig:doll_final}), not only a planar part of the bottom parallel to ground is missing, but also quite some proportion of its shape due to self-occlusion. 
Consequently, the reconstructed surface also misses these parts and features a planar surface instead.

The second case where our approach may give non-optimal results when run in an unsupervised way is on fine-detailed objects.
An example for this is, again, the baby doll which loses details on its limbs through the generalization performed by the Growing Neural Gas.
This is a known limitation of our quality metrics which are unable to determine whether a detail of the model is actually missing or there is noise present.
As generally we prefer the automatic generation of watertight models ready to use in simulation, this limitation can be accepted for our application.

\section{Conclusion}\label{sec:conclusion}
In this paper, a framework was presented which is capable of automatically generating watertight mesh representations for the use in physics simulations which, in contrary to existing approaches, works in an unsupervised way and can deal with noisy point clouds obtained from a low-cost sensor.

We have shown that our approach works on different kinds of complex objects with free-form surfaces without user interaction, even if not all sides of the object were captured by the sensor.
The major parameters of our method have been optimized in an unsupervised manner using a synthetic protoype object in order to generate visually appealing meshes which are close to the input in terms of their geometry. The resulting watertight meshes may boost fidelity in robotic simulation applications.

\bibliographystyle{IEEEtran}
\bibliography{watertight_mesh_generation}
\end{document}